\documentclass{article}
\usepackage{spconf,amsmath,graphicx}
\usepackage[draft]{fixme}
\usepackage{bm}
\usepackage[labelfont=small,font=small]{caption}
\usepackage{caption}
\usepackage{nicefrac}
\usepackage[linkcolor=blue,citecolor=blue,colorlinks=true]{hyperref}
\usepackage{multirow}
\usepackage{amssymb}
\usepackage{xcolor}

\usepackage{tikz} 
\usetikzlibrary{calc,arrows,positioning,shapes,shadows,spy,snakes,plotmarks,matrix,fit,backgrounds}

\usepackage{color}

\title{Efficient Registration of Pathological Images: \\A Joint PCA/Image-Reconstruction Approach}
\name{Xu Han$^{1}$ \qquad Xiao Yang$^{1}$ \qquad Stephen Aylward$^{2}$ \qquad Roland Kwitt$^{3}$ \qquad Marc Niethammer$^{1}$}

\address{$^{1}$University of North Carolina (UNC) at Chapel Hill, USA \\
     $^{2}$Kitware Inc., USA \\
     $^{3}$University of Salzburg, Austria}

\begin{document}
%
\maketitle
\begin{abstract}
Registration involving one or more images containing pathologies is challenging, as standard image similarity measures and spatial transforms cannot account for common changes due to pathologies. Low-rank/Sparse (LRS) decomposition removes pathologies prior to registration; however, LRS is memory-demanding and slow, which limits its use on larger data sets. Additionally, LRS blurs normal tissue regions, which may degrade registration performance.
This paper proposes an efficient alternative to LRS: (1) normal tissue appearance is captured by principal component analysis (PCA) and (2) blurring is avoided by an integrated model for pathology removal and image reconstruction. Results on synthetic and BRATS 2015 data demonstrate its utility.
\end{abstract}
%
%
\section{Introduction}
\label{section:introduction}

Image registration in the presence of pathologies is challenging as standard image similarity measures (e.g., sum of squared differences, mutual information, and normalized cross-correlation (NCC)) and standard spatial transforms (e.g., B-Spline and deformation fields) do not account for common changes arising from pathologies and cannot establish reliable spatial correspondences. Pathological image registration is needed, for example, to support (a) disease diagnosis and treatment planning using atlas-based tissue segmentation to identify traumatic brain injuries, tumors, or strokes~\cite{Irimia2012}; and (b) treatment monitoring using longitudinal images for brain tumor recurrence assessment~\cite{kwon2014}. 

A variety of approaches have been proposed to address pathological image registration. For example, cost function masking~\cite{brett2001} and geometric metamorphosis~\cite{niethammer2011} exclude pathological regions from measurements of image similarity. However, these approaches require \emph{prior segmentations} of the pathological regions, which is non-trivial and/or labor-intensive. Joint registration and segmentation approaches have also been proposed, which include estimating a latent label field to indicate missing correspondences~\cite{chitphakdithai2010,kwon2014}.

A conceptually different approach is to \emph{learn} normal image appearance from population data and to use it to estimate a quasi-normal image from an image with pathologies. This quasi-normal image can then be used for registration. Quasi-normal images can, for example, be estimated via a low-rank/sparse (LRS) decomposition~\cite{Liu2015} or by learning a direct mapping from a pathological to a quasi-normal image via an autoencoder~\cite{yang2016_autoencoder}. LRS suffers from three shortcomings: \emph{First}, the ideal LRS decomposition is computed based on already aligned images. Hence, in practice, registration and LRS decomposition steps need to be alternated making the algorithm costly. \emph{Second}, LRS decomposes the full population sample, causing high memory demand. Jointly with the first shortcoming, this severely limits the number of subjects that can be used for the decompositions to capture population variation. \emph{Third}, while LRS reconstructs pathological image areas, making them appear quasi-normal, it also blurs normal tissue and hence may impair registrations in areas unaffected by the pathology. While the autoencoder approach by Yang et al.~\cite{yang2016_autoencoder} does not blur normal tissue and does not require alternating registrations for a full population of images, it requires a large number of training images and has so far not been extended to 3D. \emph{This paper proposes an approach inspired by LRS, which overcomes all three of its shortcomings.}

\vskip0.5ex
\noindent
\textbf{Contributions.} \emph{First}, we use {\it normal} images as our population. This is different from the original LRS framework~\cite{Liu2015} which iteratively estimates quasi-normal images from groups of pathological images (interleaved with registration to a normal atlas). Instead, we can register the normal images to the atlas only \emph{once}. Additional registrations are performed only for the pathological image, greatly reducing computational cost. \emph{Second}, when LRS is applied to a population of normal images and one pathological image, the most desirable decomposition would be to allocate all normal images to the low-rank part and to decompose only the pathological image into its low-rank and sparse components\footnote{While desirable, this will not happen in practice, because part of the normal images will also be allocated to the sparse part, causing image blurring.}. Instead, we completely replace the LRS decomposition. Specifically, we mimic the low-rank component via a PCA basis obtained from the normal images in atlas space. We decompose the pathological image into (i) a quasi-normal part which is {\it close} to the PCA space and (ii) an abnormal part which has low total variation (TV) and replaces the sparse component of the LRS decomposition. This new decomposition is highly beneficial as it avoids image blurring (by only requiring \emph{closeness} to the PCA space) and captures large pathologies (via TV) while avoiding attributing image detail and misalignments to the pathology as in LRS. Similar to \cite{Liu2015}, our approach does not require prior knowledge of the location of the pathology.

\vskip0.5ex
\noindent
\textbf{Organization.} Sec.~\ref{section:methodology} discusses the LRS registration model and our proposed approach. Sec.~\ref{section:experiments_and_comparison} presents experimental results on synthetic and real data. The paper concludes with a discussion in Sec.~\ref{section:discussion} and an outline of ideas for future work.

\section{Methodology}
\label{section:methodology}
\noindent
\textbf{Review of Low-Rank/Sparse (LRS).}
The standard LRS decomposition requires minimization of
\begin{equation}
E(L,S) = \|L\|_* + \lambda\|S\|_1,\quad\text{s.t.}\quad D = L + S\enspace, 
\label{eq:lrs}
\end{equation}
where $D$ is a given data matrix, $\|\cdot\|_*$ is the nuclear norm (i.e., a convex approx. for the matrix rank), and $\lambda>0$ weighs the contribution of the sparse part, $S$, in relation to the low-rank part $L$. In imaging applications, each image is represented as a column of $D$. The low-rank term then captures common information across columns. The sparse term, on the other hand, captures uncommon/unusual information. As Eq.~\eqref{eq:lrs} is convex, minimization results in a global optimum, e.g., computed via an augmented Lagrangian approach~\cite{Lin2010}. 

To use LRS for the registration of pathological images requires joint registration and decomposition, as the decomposition relies on good spatial alignment, while good spatial alignment requires a good decomposition. This can be accomplished via alternating optimization~\cite{Liu2015}. Upon convergence, the low-rank matrix contains the normal parts of all images, while the sparse matrix contains the estimated pathologies. Since LRS does not consider spatial image information, small misalignments, unavoidable in image registration, may be considered abnormal and allocated to the sparse part. Also, image details may be allocated to the sparse part and cause blurring in the estimated normal image parts. Furthermore, solving the LRS decomposition iteratively~\cite{Lin2010} requires a singular value decomposition (SVD) at each iteration with a complexity of $\mathcal{O}(min\{mn^2,m^2n\})$\cite{holmes2007} for an $m\times n$ matrix. For large images $m\gg n$ and hence the computational complexity will grow quadratically with the number of images, $n$, making LRS costly for large sample sizes, which are beneficial to capture data variation.

\vskip1ex
\noindent
\textbf{Proposed PCA-based model.}
Our proposed model assumes that we have a collection of normal images available. In fact, our goal is to register \emph{one} pathological image to a normal-control atlas. Hence, we can first register all the normal images to the atlas using a standard image similarity measure. These normal images do not need to be re-registered during the iterative solution approach, resulting in a dramatic reduction of computational cost, which then allows using large image populations to capture normal data variation. Since we know \emph{a priori} which images are normal, we can mimic the low-rank part of LRS by a PCA decomposition of the atlas-aligned normal images;
we obtain PCA basis images $\{\beta_l\}$ and the mean image $M$. We are now only concerned with a \emph{single} pathological image $I$. Let $\hat{I}$ denote the pathological image after subtracting $M$, $B$ the PCA basis matrix, and $L$ and $S$ are images of the same size\footnote{Images are vectorized; the spatial gradient $\nabla$ is defined correspondingly.} as $I$. Our \emph{first model} minimizes
\begin{equation}
E(S, \hat{L}, \bm{\alpha}) = \gamma\|\hat{L}- B\bm{\alpha}\|_1 + \|\nabla S\|_{2,1},~\text{s.t.}~\hat{I} = \hat{L} + S
\label{eq:pca_tvl1}
\end{equation}
akin to the TV-L1 model~\cite{chan2005}, where $\|\nabla S\|_{2,1}=\sum_i \|\nabla S_i\|_2$ and $i$ denotes spatial location. The \emph{second model} minimizes
\begin{equation}
\label{eq:pca_rof}
E(S, \hat{L}, \bm{\alpha}) = \frac{\gamma}{2}\|\hat{L}-B\bm{\alpha}\|_2^2 + \|\nabla S\|_{2,1},~\text{s.t.}~\hat{I} = \hat{L} + S
\end{equation}
and is akin to the Rudin-Osher-Fatemi (ROF) model~\cite{rudin1992}. Both models result in a TV term, $S$, which explains the parts of $\hat{I}$ which are (i) spatially contiguous, (ii) relatively large, and (iii) cannot be explained by the PCA basis, e.g., a tumor region. The quasi-low-rank part $\hat{L}$ remains close to the PCA space but retains fine image detail. Adding $M$ to $\hat{L}$ results in the reconstructed quasi-normal image ${L}$. In principle, model~\eqref{eq:pca_tvl1} would be preferred, because of the attractive geometric scale-space properties of the TV-L1 model~\cite{chan2005}. However, we use model~\eqref{eq:pca_rof} in our experiments as it is simpler to optimize. Unfortunately, just as the ROF model~\cite{chan2005}, it suffers from an intensity loss. We can counteract this effect by adapting the iterative regularization approach proposed by Osher et al.~\cite{osher2005} for the ROF model, which iteratively adds ``noise'' back to the original images. Specifically, we first solve~\eqref{eq:pca_rof} (obtaining $\tilde{L}_0=\hat{L}$ and $\bm{\alpha}_0$), followed by a small number of regularization steps. For each iteration $k\geq 1$, we minimize
%
%
\begin{equation}
\label{eq:rof_reg}
\begin{split}
E(S_{k}, \tilde{L}_{k}, \bm{\alpha}_{k}) &= \frac{\gamma}{2}\|\tilde{L}_{k}-B\bm{\alpha}_{k}\|_2^2~+
\|\nabla S_{k}\|_{2,1}, \\~\text{s.t.}~\hat{I}_{k} &=\tilde{L}_{k} + S_{k}\enspace,
\end{split}
\end{equation}
where $\hat{I}_k = \hat{I} + \tilde{L}_{k-1} - B\bm{\alpha}_{k-1}$. After $N$ iterations, the TV part, $S_N$, will contain an approximation of the pathology, from which we obtain the quasi-normal image $\hat{L}_N = \hat{I}-S_N$. The quasi-normal image reconstructs pathological areas while retaining detailed image information in normal image areas. 


\vskip0.5ex
\noindent
\textbf{Implementation details.} We solve our PCA model via a primal-dual hybrid gradient method \cite{goldstein2013}. Compared to LRS, \emph{memory requirements are lower and runtime is faster.}\\[-8mm]

\section{Experiments}
\label{section:experiments_and_comparison}

We use the ICBM atlas~\cite{fonov2009} as our normal atlas image.

\textbf{Quasi-tumor data (2D).} We evaluate the performance of our model in 2D. We pick 250 images from the OASIS cross-sectional MRI dataset~\cite{marcus2010} as the normal population. We simulate 50 test cases by picking another set of 50 OASIS images and registering them to the BRATS 2015 T1c images~\cite{menze2015}, followed by injecting the BRATS tumors into these warped images. The registrations simulate tumor mass effects. Each image is of size 196$\times$232 with $1$mm isotropic pixels. We select 50 fixed normal images as the population for LRS, to test a scenario which would still be computable in 3D given the high computational demand of LRS. We select 250 normal images for our PCA model and choose the top 150 PCA modes as the PCA basis. We test \emph{without} regularization and \emph{with} at most two regularization steps.
\begin{table}[htb]
  \centering
  \setlength\tabcolsep{-0.0125\columnwidth}
	\begin{tabular}{ccccc}
          \begin{tikzpicture}[thick, spy using outlines={rectangle,lens={scale=2}, size=1.0cm, connect spies}]
	    \node{\includegraphics[width=0.2\columnwidth]{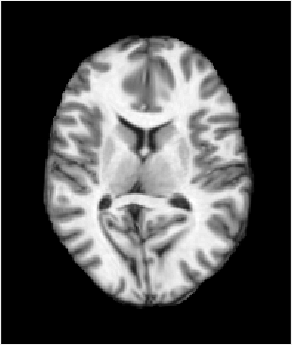}};%
            \spy [green] on (0.2,0.55) in node [left] at (0.8,1.4);%
          \end{tikzpicture}
          &
          \begin{tikzpicture}[thick, spy using outlines={rectangle,lens={scale=2}, size=1.0cm, connect spies}]
	    \node {\includegraphics[width=0.2\columnwidth]{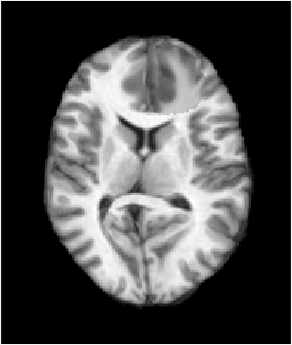}};
            \spy [green] on (0.2,0.55) in node [left] at (0.8,1.4);
          \end{tikzpicture}
          &
          \begin{tikzpicture}[thick, spy using outlines={rectangle,lens={scale=2}, size=1.0cm, connect spies}]
	    \node {\includegraphics[width=0.2\columnwidth]{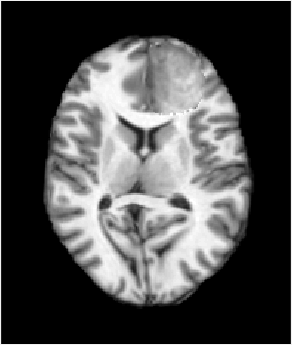}};
            \spy [green] on (0.2,0.55) in node [left] at (0.8,1.4);
          \end{tikzpicture}
            &
            \begin{tikzpicture}[thick, spy using outlines={rectangle,lens={scale=2}, size=1.0cm, connect spies}]
	      \node {\includegraphics[width=0.2\columnwidth]{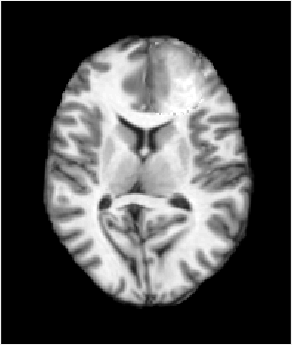}};
              \spy [green] on (0.2,0.55) in node [left] at (0.8,1.4);
            \end{tikzpicture}
              &
              \begin{tikzpicture}[thick, spy using outlines={rectangle,lens={scale=2}, size=1.0cm, connect spies}]
	        \node {\includegraphics[width=0.2\columnwidth]{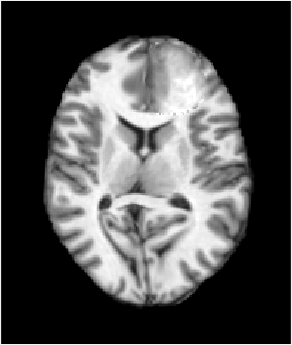}};
                \spy [green] on (0.2,0.55) in node [left] at (0.8,1.4);
              \end{tikzpicture}
                \\[-2mm]
	  \includegraphics[width=0.2\columnwidth]{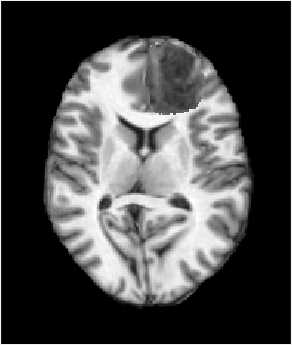} &
	  \includegraphics[width=0.2\columnwidth]{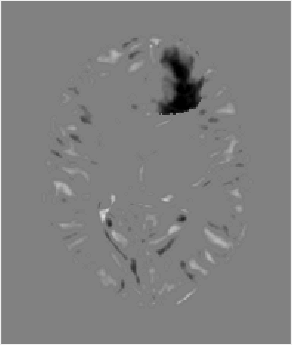} &
	  \includegraphics[width=0.2\columnwidth]{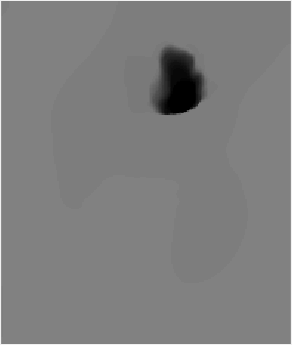} &
	  \includegraphics[width=0.2\columnwidth]{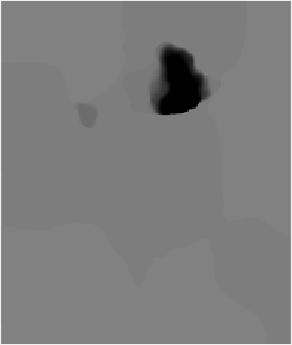} &
	  \includegraphics[width=0.2\columnwidth]{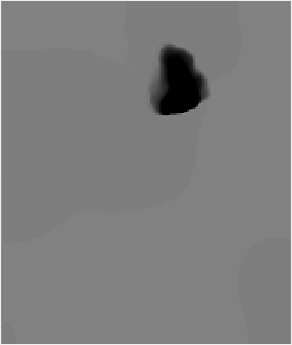} \\
	  \small (a)&\small(b)&\small(c)&\small(d)&\small(e)\\[-3mm]
    \end{tabular}
	\captionof{figure}{Example quasi-normal reconstructions. (a) ground truth (\emph{top}) and tumor (\emph{bottom}); (b)-(e) Reconstruction result (\emph{top}) and tumor (\emph{bottom}): (b) LRS; (c) PCA model w/o regularization; (d) PCA model w/ one and (e) w/ two regularization steps.}
	\label{fig:reconstructions}
\end{table}

Fig.~\ref{fig:reconstructions} shows reconstruction results for LRS and for our PCA-based models. For each model, we perform cross-validation, partitioning the 50 test cases into 10 folds, with 9 folds for training and 1 fold for testing. We train each model with $\lambda = \{$0.005, 0.0067, 0.0084, 0.01, 0.0117, 0.0133, 0.015$\}$, for LRS, and $\gamma=\{$0.5, 1, 1.5, 2, 2.5,3$\}$, for our PCA models. We evaluate the \emph{mean registration error} compared to the ground truth registration result. This is done in three areas: the tumor area, the normal areas near the tumor (within 10mm) and the normal areas far from the tumor ($>$10mm). We weigh the deformation errors in these areas 4:1:1 and, for each model, pick the parameter that gives the smallest errors.

Fig.~\ref{fig:reconstructions} shows a good but blurry LRS reconstruction as the sparse part captures the tumor \emph{and} misalignments. Also, the small and round left posterior ventricle in the ground truth image is not reconstructed faithfully by LRS. Our PCA models capture only the tumor in $S$, resulting in a sharper and more precise reconstruction. Furthermore, regularization yields an even better tumor separation.
\begin{table}[htb]
  \centering
    \setlength\tabcolsep{-0.0125\columnwidth}
	\begin{tabular}{cccccc}
          \begin{tikzpicture}[thick, spy using outlines={rectangle,lens={scale=2}, size=1.0cm, connect spies}]
	    \node{\includegraphics[width=0.165\columnwidth]{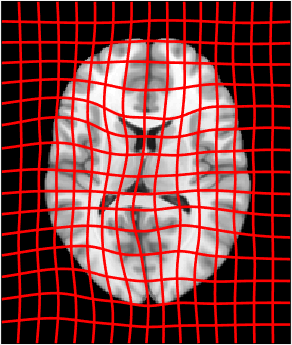}};
            \spy [blue, every spy on node/.append style={thick}] on (0,0.4) in node [left] at (0.6,1.3);
          \end{tikzpicture}&
          \begin{tikzpicture}[thick, spy using outlines={rectangle,lens={scale=2}, size=1.0cm, connect spies}]
	    \node{\includegraphics[width=0.165\columnwidth]{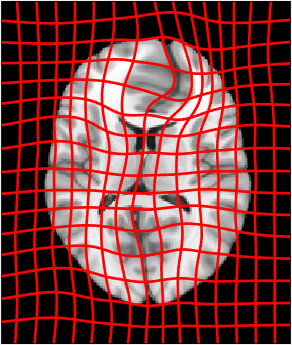}};
            \spy [blue, every spy on node/.append style={thick}] on (0,0.4) in node [left] at (0.6,1.3);
          \end{tikzpicture}&
          \begin{tikzpicture}[thick, spy using outlines={rectangle,lens={scale=2}, size=1.0cm, connect spies}]
	    \node{\includegraphics[width=0.165\columnwidth]{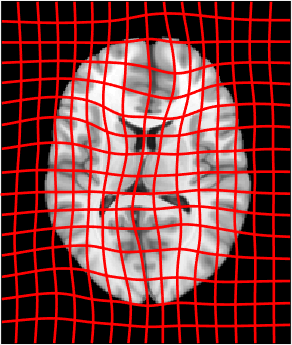}};
            \spy [blue, every spy on node/.append style={thick}] on (0,0.4) in node [left] at (0.6,1.3);
          \end{tikzpicture}&
          \begin{tikzpicture}[thick, spy using outlines={rectangle,lens={scale=2}, size=1.0cm, connect spies}]
	    \node{\includegraphics[width=0.165\columnwidth]{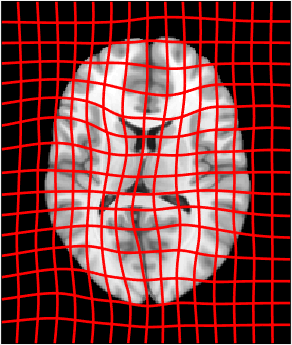}};
            \spy [blue, every spy on node/.append style={thick}] on (0,0.4) in node [left] at (0.6,1.3);
          \end{tikzpicture}&
          \begin{tikzpicture}[thick, spy using outlines={rectangle,lens={scale=2}, size=1.0cm, connect spies}]
	    \node{\includegraphics[width=0.165\columnwidth]{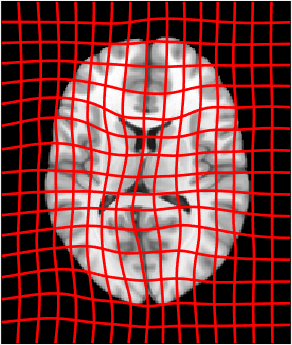}};
            \spy [blue, every spy on node/.append style={thick}] on (0,0.4) in node [left] at (0.6,1.3);
          \end{tikzpicture}&
          \begin{tikzpicture}[thick, spy using outlines={rectangle,lens={scale=2}, size=1.0cm, connect spies}]
	    \node{\includegraphics[width=0.165\columnwidth]{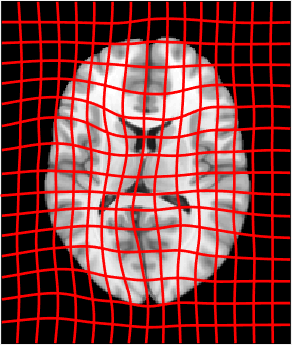}};
            \spy [blue, every spy on node/.append style={thick}] on (0,0.4) in node [left] at (0.6,1.3);
          \end{tikzpicture}\\[-1mm]
		\small(a)&\small(b)&\small(c)&\small(d)&\small(e)&\small(f)\\[-3mm]
	\end{tabular}
	\captionof{figure}{Example atlas-to-image registrations: (a) ground truth; (b) tumor; (c) LRS; (d) PCA model w/o regularization; (e) PCA model w/ one step and (f) w/ two regularization steps.}
	\label{fig:registrations}
\end{table}
\begin{table}[htb]
	\centering
	\begin{tabular}{@{}c@{}c@{}c@{}c@{}c@{}}
			\includegraphics[width=0.2\columnwidth]{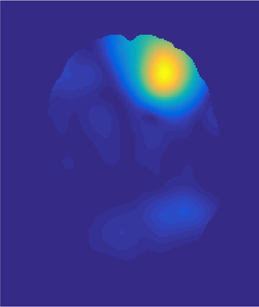} &
			\includegraphics[width=0.2\columnwidth]{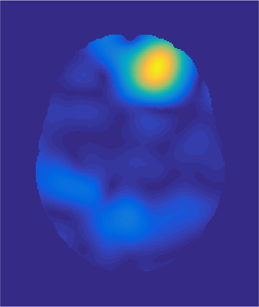} &
			\includegraphics[width=0.2\columnwidth]{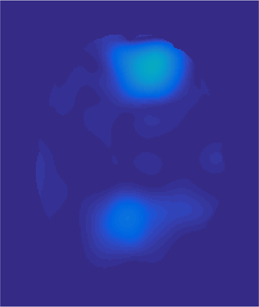} &
			\includegraphics[width=0.2\columnwidth]{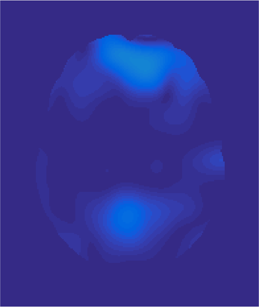} &
			\includegraphics[width=0.2\columnwidth]{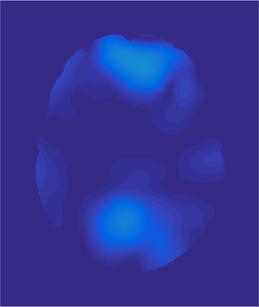} \\[-1mm]
			\includegraphics[width=0.2\columnwidth]{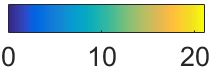} &
				\multicolumn{4}{@{}r@{}}{\includegraphics[width=0.78\columnwidth]{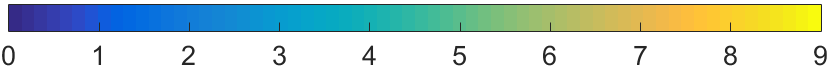}}\\
		\small(a)&\small(b)&\small(c)&\small(d)&\small(e)\\[-3mm]
	\end{tabular}
	\captionof{figure}{Example atlas-to-image registration errors [mm]: (a) tumor; (b) LRS; (c) PCA model w/o regularization; (d) PCA model w/ one step of regularization; (e) PCA model w/ two steps of regularization.}
	\label{fig:registrations-error}
\end{table}

Fig.~\ref{fig:registrations} shows atlas-to-image registration results for images with and without tumor, LRS reconstruction and our PCA-based models with and without regularization. Fig.~\ref{fig:registrations-error} shows the spatial error distributions, compared to the ground truth registration. We use \texttt{NiftyReg}~\cite{modat2010} (with standard settings) and NCC for registrations. Errors are computed using Euclidean distance. Direct registration of the tumor image results in large registration errors. Registration to the low-rank reconstruction greatly reduces the error in the tumor areas but retains errors near the cortex, mainly due to its blurry reconstruction. Our PCA models further reduce registration errors in the tumor areas \emph{and} keep errors near the cortex low.
\begin{table}[htb]
	\centering
	\begin{tabular}{@{}c@{}c@{}c@{}}
			\includegraphics[width=0.33\columnwidth]{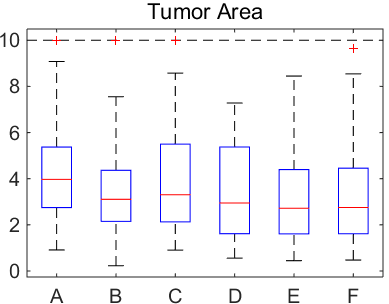} &
			\includegraphics[width=0.33\columnwidth]{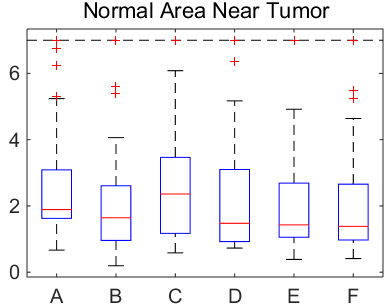} &
			\includegraphics[width=0.33\columnwidth]{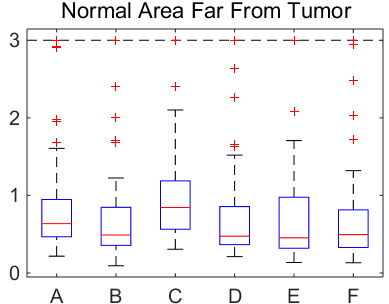}\\[-3mm]
	\end{tabular}
	\captionof{figure}{Mean deformation errors [mm] for test cases. \emph{A:} Tumor image; \emph{B:} cost function masking; \emph{C:} LRS; \emph{D:} PCA model w/o regularization; \emph{E:} PCA w/ one and \emph{F:} w/ two regularization steps.}
	\label{fig:errors}
\end{table}

Fig.~\ref{fig:errors} shows mean deformation errors over all test cases in the 3 areas. We also add cost function masking for comparison. Note that the tumors selected from BRATS to generate our 2D test cases are relatively mild resulting in relatively small deformation errors even when using tumor images for registration. LRS (C) reduces errors in the tumor areas but has higher errors in the normal areas. Our PCA models (D, E, F) show better results in both the tumor and the normal areas. Paired $t$-tests between LRS and our PCA models show statistically significant differences in all areas for the PCA models with regularization, and in the normal areas for the PCA model without regularization. Moreover, the PCA models with regularization show similar performance to cost function masking but do not require a tumor segmentation.

\vskip0.5ex
\noindent
\textbf{Quasi-tumor Dataset (3D).} We also generate 3D quasi-tumor data for evaluation. We pick 100 OASIS images and select the first 50 PCA modes as the basis. We also simulate 20 test images with tumor. Each image is of size 196$\times$232$\times$189. Different from the 2D experiment, the tumors for our 3D test cases are picked randomly from BRATS, including cases with large tumors and deformations.
\begin{table}[htb]
	\centering
	\begin{tabular}{@{}c@{}c@{}c@{}}
			\includegraphics[width=0.33\columnwidth]{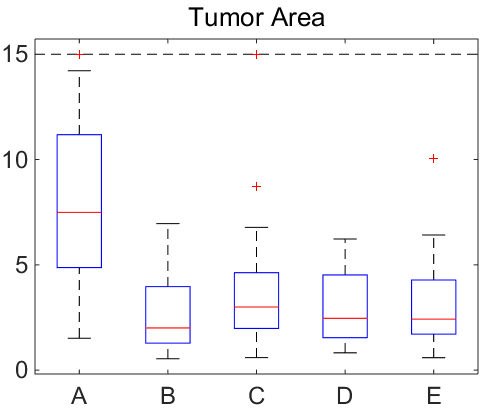} &
			\includegraphics[width=0.33\columnwidth]{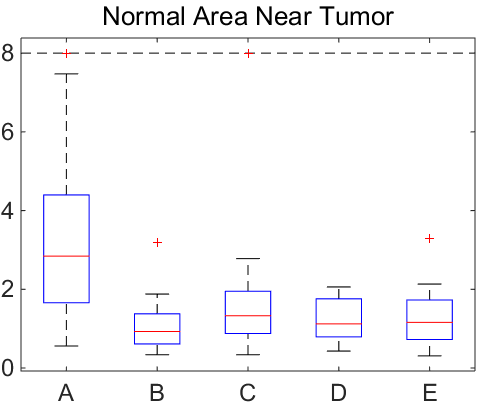} &
			\includegraphics[width=0.33\columnwidth]{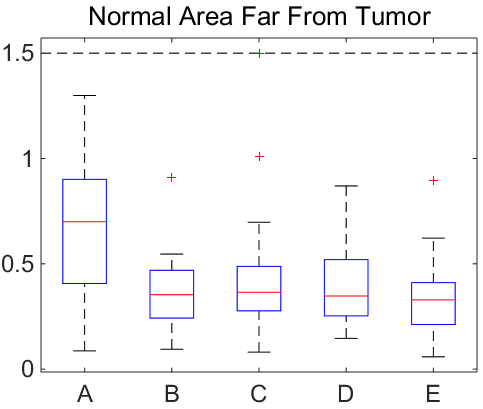}\\[-3mm]
	\end{tabular}
	\captionof{figure}{Mean deformation errors [mm] for 3D test cases. \emph{A:} Tumor image; \emph{B:} cost function masking; \emph{C:} PCA model w/o regularization; \emph{D:} PCA model w/ one and \emph{E:} w/ two regularization steps.}
	\label{fig:errors-3d}
\end{table}
For cross validation, we separate the twenty test cases into ten 9:1 folds. The training parameters for our models are $\gamma=\{1$, 1.5, 2, 2.5, 3$\}$. Registration errors in the three different areas are weighted as before, i.e., 4:1:1.
Fig.~\ref{fig:errors-3d} shows box plots of the mean deformation errors. Directly registering to tumor images results in large errors. The quasi-normal images reconstructed by our PCA models greatly reduce the deformation errors in all the areas. As in 2D, our PCA models show similar performance to cost function masking but do not require a tumor segmentation.

\vskip0.5ex
\noindent
\textbf{BRATS Dataset (3D).}
Finally, we also apply our model to the BRATS 2015 data~\cite{menze2015}. As the BRATS data was acquired at different institutions and on different scanners, we pick 80 BRATS T1c images as the population which show consistent image appearance and contain the full brain. To obtain our PCA model we locally impute image intensities in the tumor areas, prior to computing the PCA basis, using the mean intensity over all images that do not contain a tumor at that location. We also pick the first 50 PCA modes as our basis.

\begin{table}[htb]
	\centering
	\begin{tabular}{@{}c@{}c@{}c@{}c@{}}
		\multirow{2}{*}[2em]{
			\includegraphics[width=0.2\columnwidth]{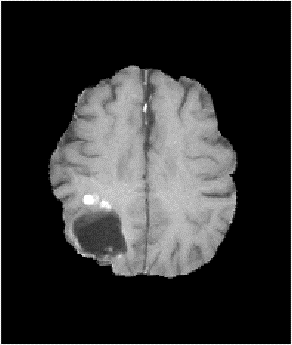}
	    }&
			\includegraphics[width=0.2\columnwidth]{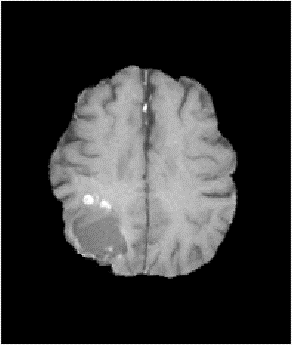} &
			\includegraphics[width=0.2\columnwidth]{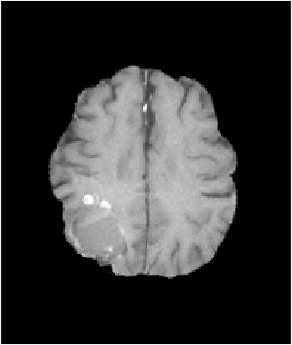} &
			\includegraphics[width=0.2\columnwidth]{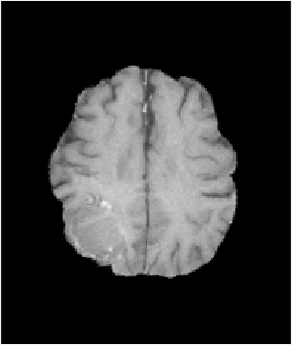}\\
                        &
			\includegraphics[width=0.2\columnwidth]{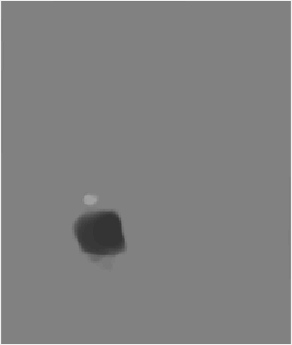} &
			\includegraphics[width=0.2\columnwidth]{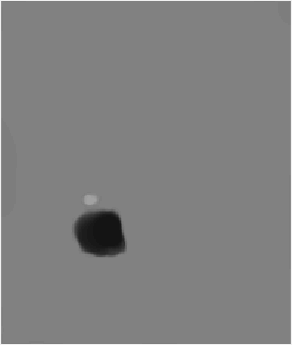} &
			\includegraphics[width=0.2\columnwidth]{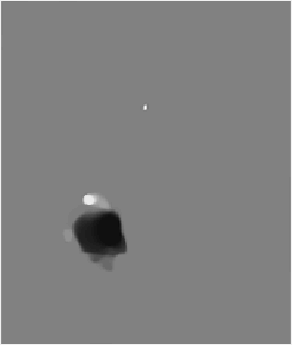} \\
		\small(a)&\small(b)&\small(c)&\small(d)\\[-3mm]
	\end{tabular}
	\captionof{figure}{Example BRATS reconstructions: (a) tumor image; (b)-(d) reconstructions (\emph{top}) and tumors (\emph{bottom}); (b) PCA model w/o regularization; (c) PCA model w/ one and (d) w/ two regularization steps.}
	\label{fig:recontructions-brats}
\end{table}
Fig.~\ref{fig:recontructions-brats} shows decomposition results for our PCA models. We pick $\gamma=5.0$ for the model without and $\gamma=2.0$ for models with regularization. The goal is to allocate as much of the tumor as possible to the abnormal part, $S$, while keeping the normal tissue in the quasi-normal part of the decomposition. Qualitatively, our models identify tumor/normal areas, while retaining image details in normal tissue areas.
\begin{table}[htb]
  \centering
  \setlength\tabcolsep{-0.0125\columnwidth}
  \begin{tabular}{ccccc}
    \begin{tikzpicture}[thick, spy using outlines={rectangle,lens={scale=2}, size=1.0cm, connect spies}]
      \node{\includegraphics[width=0.2\columnwidth]{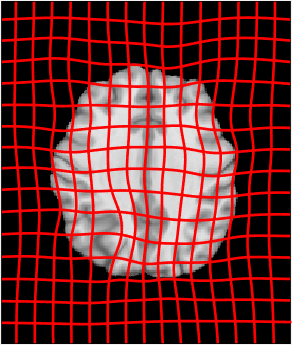}};
      \spy [blue, every spy on node/.append style={thick}] on (-0.275,-0.35) in node [left] at (0.6,-1.3);
    \end{tikzpicture}&
    \begin{tikzpicture}[thick, spy using outlines={rectangle,lens={scale=2}, size=1.0cm, connect spies}]
      \node{\includegraphics[width=0.2\columnwidth]{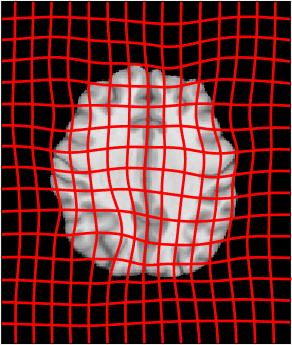}};
      \spy [blue, every spy on node/.append style={thick}] on (-0.275,-0.35) in node [left] at (0.6,-1.3);
    \end{tikzpicture}&
    \begin{tikzpicture}[thick, spy using outlines={rectangle,lens={scale=2}, size=1.0cm, connect spies}]
      \node{\includegraphics[width=0.2\columnwidth]{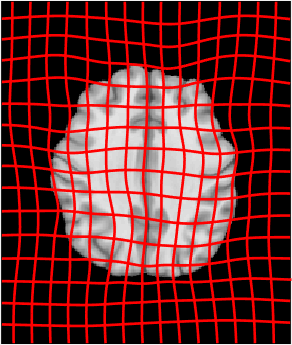}};
      \spy [blue, every spy on node/.append style={thick}] on (-0.275,-0.35) in node [left] at (0.6,-1.3);
    \end{tikzpicture}&
    \begin{tikzpicture}[thick, spy using outlines={rectangle,lens={scale=2}, size=1.0cm, connect spies}]
      \node{\includegraphics[width=0.2\columnwidth]{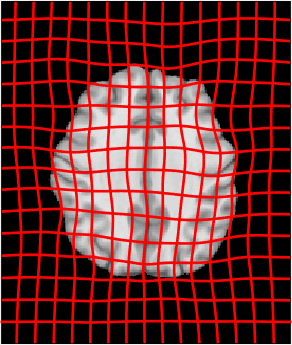}};
      \spy [blue, every spy on node/.append style={thick}] on (-0.275,-0.35) in node [left] at (0.6,-1.3);
    \end{tikzpicture}&
    \begin{tikzpicture}[thick, spy using outlines={rectangle,lens={scale=2}, size=1.0cm, connect spies}]
      \node{\includegraphics[width=0.2\columnwidth]{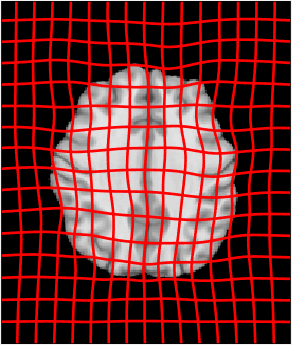}};
      \spy [blue, every spy on node/.append style={thick}] on (-0.275,-0.35) in node [left] at (0.6,-1.3);
    \end{tikzpicture}\\
    \small(a)&\small(b)&\small(c)&\small(d)&\small(e)\\[-3mm]
  \end{tabular}
  \captionof{figure}{Example BRATS atlas-to-image registration results: (a) tumor image; (b) cost function masking; (c) PCA model w/o regularization; (d) PCA model w/ one and (e) w/ two regularization steps.}
  \label{fig:registrations-brats}
  \vspace{-0.2cm}
\end{table}
Finally, Fig.~\ref{fig:registrations-brats} shows atlas-to-image registration results for the PCA models, the tumor image, and cost function masking. The results show the significant impact of the tumor on the registration, which is mitigated by cost function masking and our PCA models, in particular, with regularization.

\vskip0.5ex
\noindent
\textbf{Memory use.} For LRS, $D\in\mathbb{R}^{m\times n}$, where $m$ is the number of pixels/voxels and $n$ the number of images. Each 196$\times$232$\times$189 3D image (stored as double) consumes about 65MB of memory. Hence, 3GB of memory is required to store $D$ for $n=50$. As the LRS algorithm~\cite{Lin2010} requires storing several variables of the size of $D$, memory use quickly becomes prohibitive, in particular for GPU implementations. Our model only stores \emph{one} copy of the pre-computed PCA basis thereby substantially reducing memory use ($\approx$ 4GB/8GB for $n=50$ in single/double precision) and consequentially facilitating larger sample sizes even on the GPU.

\vskip0.5ex
\noindent
\textbf{Runtime.} For the 3D cases, with $n=50$, an LRS decomposition takes one hour to run and uses up to 40GB of memory thereby precluding a GPU implementation. Due to the low memory requirements of our PCA models, a GPU implementation is possible resulting in a runtime of $\approx$3 minutes / decomposition. The 3D image registrations are computed on the CPU ($\approx$3 minutes). Therefore, with 6 registration iterations, our algorithm requires $\approx$40 minutes / test case and takes about 1 hour if extra regularization steps are computed, whereas the LRS approach takes $>$6 hours.\\[-7mm]


\section{Discussion}
\label{section:discussion}

To conclude, our experiments show that the proposed PCA-based model (i) improves image reconstructions and registrations over the LRS model, while (ii) requiring less memory, at (iii) substantially reduced computational cost. On the tested quasi-tumor data, our models achieve performance close to cost function masking, without requiring tumor segmentations. Future work should include a quantitative assessment of the registration results on 3D BRATS data via landmarks. 

This research is supported by NIH 1 R41 NS091792-01, NSF EECS-1148870 and NIGMS/NIBIB:R01EB021396.

\let\oldbibliography\thebibliography
\renewcommand{\thebibliography}[1]{%
  \oldbibliography{#1}%
  \setlength{\itemsep}{0pt}%
}

\small
\bibliographystyle{IEEEbib}
\bibliography{strings,refs}

\end{document}